%% file: sample-sigconf.tex
\documentclass[sigconf]{acmart}

\newtheorem{defn}{Definition}[section]

\usepackage{graphicx} 
\usepackage{multirow}
\usepackage{multicol}
\usepackage{rotating}
\usepackage{amsmath}
\usepackage{amsthm}
\usepackage{amsfonts}
\usepackage{booktabs}
\usepackage{verbatim}
\usepackage{subfigure}
\usepackage{algorithm}  
\usepackage{algorithmic}
\usepackage{array}

\usepackage{url}
\usepackage{xspace}
\usepackage{enumitem}

\AtBeginDocument{%
  \providecommand\BibTeX{{%
    \normalfont B\kern-0.5em{\scshape i\kern-0.25em b}\kern-0.8em\TeX}}}

\copyrightyear{2021} 
\acmYear{2021} 
\setcopyright{acmlicensed}\acmConference[IJCKG'21]{The 10th International Joint Conference on Knowledge Graphs}{December 6--8, 2021}{Virtual Event, Thailand}
\acmBooktitle{The 10th International Joint Conference on Knowledge Graphs (IJCKG'21), December 6--8, 2021, Virtual Event, Thailand}
\begin{document}

% \fancyhead{}

\title{Knowledge Graph Embedding in E-commerce Applications: Attentive Reasoning, Explanations, and Transferable Rules}

\author{Wen Zhang}
\email{wenzhang2015@zju.edu.cn}
\affiliation{%
\institution{Zhejiang University}
% \country{China}
\country{}
}

\author{Shumin Deng}
\email{231sm@zju.edu.cn}
\affiliation{%
\institution{Zhejiang University}
% \country{China}
\country{}
}

\author{Mingyang Chen}
\email{mingyangchen@zju.edu.cn}
\affiliation{%
\institution{Zhejiang University}
% \country{China}
\country{}
}

\author{Liang Wang}
\email{liangwang.wl@alibaba-inc.com}
\affiliation{%
\institution{Alibaba Group}
% \country{China}
\country{}
}

\author{Qiang Chen}
\email{lapu.cq@alibaba-inc.com}
\affiliation{%
\institution{Alibaba Group}
% \country{China}
\country{}
}

\author{Feiyu Xiong}
\email{feiyu.xfy@alibaba-inc.com}
\affiliation{%
\institution{Alibaba Group}
% \country{China}
\country{}
}

\author{Xiangwen Liu}
\email{vicki.liux@alibaba-inc.com}
\affiliation{%
\institution{Alibaba Group}
\country{}
}

\author{Huajun Chen}
\authornote{Corresponding author}
\email{huajunsir@zju.edu.cn}
\affiliation{%
\institution{Zhejiang University \& AZFT Joint Lab for Knowledge Engine}
\country{}
% \country{China}
}

% \renewcommand{\shortauthors}{ Zhang and Deng et al.}

\begin{abstract}
Knowledge Graphs (KGs), representing facts as triples, have been widely adopted in many applications. Reasoning tasks such as link prediction and rule induction are important for the development of KGs. Knowledge Graph Embeddings (KGEs) embedding entities and relations of a KG into continuous vector spaces, have been proposed for these reasoning tasks and proven to be efficient and robust. But the plausibility and feasibility of applying and deploying KGEs in real-work applications has not been well-explored. In this paper, we discuss and report our experiences of deploying KGEs in a real domain application: e-commerce. We first identity three important desiderata for e-commerce KG systems: 1)  attentive reasoning, reasoning over a few  target relations of more concerns instead of all; 2) explanation, providing explanations for a prediction to help both users and business operators understand why the prediction is made; 3) transferable rules, generating reusable rules to accelerate the deployment of a KG to new systems. While non existing KGE could meet all these desiderata, we propose a novel one, an explainable knowledge graph attention network that make prediction through modeling correlations between triples rather than purely relying on its head entity, relation and tail entity embeddings. It could automatically selects attentive triples for prediction and records the contribution of them at the same time, from which explanations could be easily provided and transferable rules could be efficiently produced. We empirically show that our method is capable of meeting all three desiderata in our e-commerce application and outperform typical baselines on datasets from real domain applications. 
\end{abstract}

\begin{CCSXML}
<ccs2012>
<concept>
<concept_id>10010147.10010178.10010187</concept_id>
<concept_desc>Computing methodologies~Knowledge representation and reasoning</concept_desc>
<concept_significance>500</concept_significance>
</concept>
<concept>
<concept_id>10010147.10010257.10010293.10010314</concept_id>
<concept_desc>Computing methodologies~Rule learning</concept_desc>
<concept_significance>500</concept_significance>
</concept>
</ccs2012>
\end{CCSXML}

\ccsdesc[500]{Computing methodologies~Knowledge representation and reasoning}
\ccsdesc[500]{Computing methodologies~Rule learning}

\keywords{Knowledge Graphs; Representation Learning; Reasoning, E-commerce; Explainable AI; Rules}


\maketitle

\input{introduction}

\input{preliminary}

\input{method}

\input{experiment}

\input{related-work}

\section{Conclusion and future work}
In this paper, we discuss the implementation of KG embedding methods in e-commerce application. 
We identify that, in real-life applications, attentive reasoning, explanations and transferable rules are necessary, because real-life applications are usually task-specific and human-machine collaborated scenarios. 
We propose a new KGE method, explainable knowledge graph attention network, which predict a triple based on its neighbor triples via an attentive mechanism, enabling providing explanations based on attention weights. And we propose to generate rules from explanations. It proves to be friendly for rule quality and diversity.
In the future, we would like to adapt our method to more knowledge graph embedding methods and apply them to other attentive reasoning tasks.

\begin{acks}
This work is funded by NSFCU19B2027/91846204, national key
research program 2018YFB1402800.
\end{acks}

\bibliographystyle{ACM-Reference-Format}
\bibliography{sample-sigconf.bib}


\end{document}

%% file: introduction.tex
\section{Introduction}
Knowledge Graphs (KGs) represent facts as triples, such as \emph{(IPhone, brandIs, Apple)}. 
Due to the convenience of fusing data from various sources and building semantic connections between entities, 
KGs have been widely applied in industry and many huge KG have been built (e.g. Google's Knowledge Graph, Facebook's Social Graph, and Alibaba/Amazon's e-commerce Knowledge Graph).

\begin{figure*}[htbp] 
\centering 
	\includegraphics[scale=0.30]{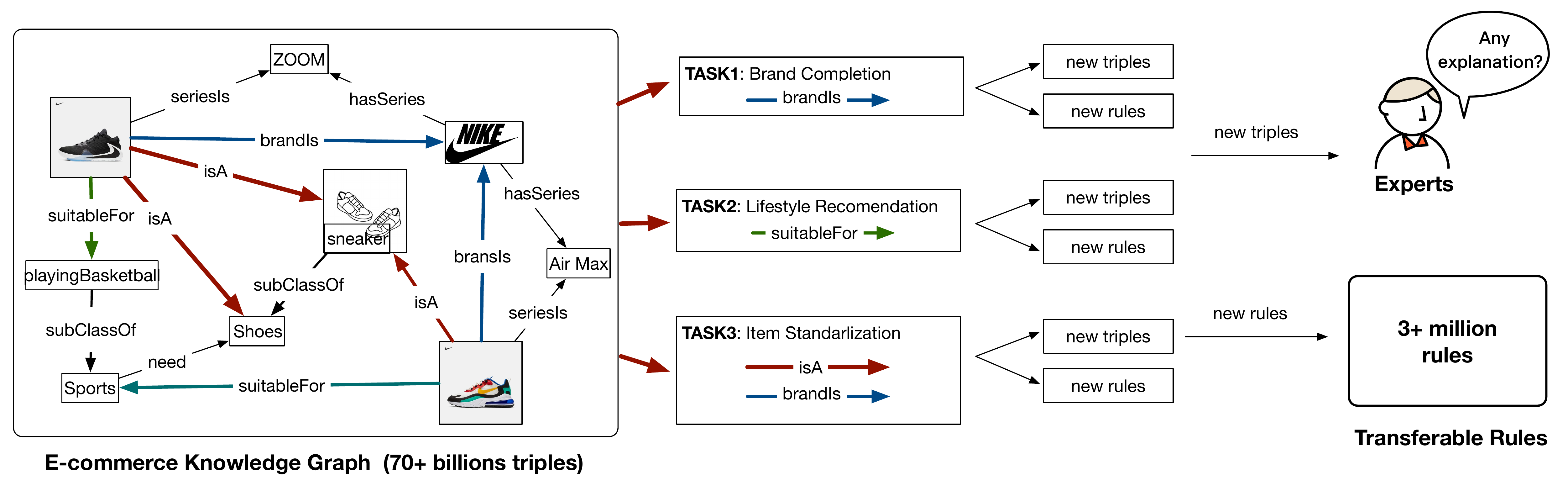}
	\vspace{-3mm}
	\caption{Overview of knowledge graph reasoning system in e-commerce application.}
	\label{motivation}
	\vspace{-5mm}
\end{figure*}

In this paper, we report our experiences on e-commerce knowledge graph application. 
Our e-commerce knowledge graph~\cite{DBLP:conf/icde/ZhangWYWZC21} is built as a union way to integrate massive information about items, online-shops, users, etc., on the e-commerce platform. 
It help us build the integrated view of data and contribute to a variety of business tasks ranging from searching, question answering, recommendation~\cite{DBLP:conf/icde/WongFZVCZHCZC21} and business intelligence, etc. 
Currently it contains 70+ billion triples and 3+ million rules. Being similar to other KGs, it is still far from complete, while different to other KGs, we typically perform task-specific completion for the e-Commerce KG. 
As Figure~\ref{motivation} illustrates, for one specific business task such as brand completion and item recommendation, we usually pay attention on only a few \emph{target relations}. For each task, new triples and rules related to target relations will be generated. Then new triples will be presented to business operators to check the correctness, and rules will be added into the transferable pool for future reuse. Thus our e-commerce knowledge graph application is a task specific and human-machine collaborated scenario. Usually new triples and rules are inferred by predicting algorithm and rule learning method independently, most of which are \emph{black-box models}. In order to avoid tedious algorithm design and enforce the efficiency of the human-machine collaboration cycle, we propose three desiderata for the algorithms applied to e-commerce knowledge graph applications: 
1) \textbf{\emph{Attentive reasoning}}: reasoning over a few attentive relations concerned to specific tasks instead of all relations;
2) \textbf{\emph{Explanations}}: providing explanations for predicted results to both end users and business operators;
3) \textbf{\emph{Transferable rules}}: generating reusable rules easily transferable to new tasks or new systems.

Recently considerable \emph{Knowledge Graph Embedding (KGE)} methods~\cite{TransE:conf/nips/BordesUGWY13, DistMul:conf/iclr/2015, ConvE:conf/aaai/DettmersMS018, ANALOGY:conf/icml/LiuWY17,DBLP:conf/nlpcc/ZhangLC18, DBLP:conf/jist/ZhangDWCZC19}, have been proposed for KG applications. They embed knowledge graph into continuous vector space and calculate the truth value of a triple via entity and relation embeddings. As KGE methods prove to be effective, we also consider to apply them, while as far as we know, non existing KGE method could meet the three desiderata together in our scenario. 
Firstly, conventional KGE models usually take all relations and triples into training equally which are unnecessarily time consuming and will decrease the performance for those concerned or attentive target relations, thus not meeting the attentive reasoning desiderata.
Secondly, they inject all entity/relation features into compact and low dimension vectors and focus on encoding individual triples, thus no internal explanation could be provided to indicate how the results are derived and predicted. Thirdly, conventional models usually produce embeddings for only entities and relations, while in many e-commerce applications, rules play a pivotal roles in modeling business knowledge and transferable rules are highly valuable in accelerating the deployment of a KG to new systems.

Thus is the work, we propose our idea of tackle these challenges. To adapt to attentive reasoning task, an intuitive idea is to train with only triples containing target relations, while this will disassembles an densely linked KG and lose too much information. Thus the method should choose necessary and related triples, so that the best prediction performance for target relations could be guaranteed. 
For explanations, we represent them as a set of other triples for a prediction, thus the method should establish association among other triples and the predicted one, from which explanations can be easily derived. 
For rules, generating them from explanations could be an alternative way, as explanations could be regarded as groundings of rules. With these observations and analysis, we emphasize that \emph{the key to addressing the three challenges is to properly model the correlations among triples.}

We then propose a novel KGE method, an explainable knowledge graph attention network built based on translation assumption from TransE~\cite{TransE:conf/nips/BordesUGWY13}.
To model correlations between triples, it predicts the truth value of a triple $(h,r,t)$ based on its neighbor triples rather than purely rely on the embedding of $h, r$ and $t$. 
We devise a basic layer to aggregate triples' one-degree neighbor triple information to head entity via a graph attention network. The basic layer could be stacked to enable the model aware of higher-degree triples. 
We keep track of the \emph{attentive weight} of each neighbor triple in each layer and generate explanations via multiplication between these weights and select top $k$ weighted ones. 
We then generate rules from explanations via replacing shared entities as variables and select quality ones generated multiple times.  

In the experiment with a real e-commerce knowledge graph dataset, we firstly present results on partial link prediction, as a task for attentive reasoning. Then we evaluate the quality of explanations provided by algorithms via statistic metrics and how it help people check the correctness of predicted triples. And finally we compare the quality of rules generated from our method with baselines in terms of quality rules, and the number of new triples they infer. Experiment results show that our method successfully meets the desiderata, and is data efficient.

% main contributions
Our main contributions are as follows:
\begin{itemize}
	\item We report a real-life e-commerce KGE application and state that desired characters of algorithms applied in this scenario include conducting attentive reasoning, providing explanations for prediction and generating transferable rules.
	\item We  propose a novel KGE method, the explainable knowledge graph attention network  building correlations between triples via a graph attention mechanism and proves to be effective in addressing the challenges in a unified framework. 
	\item We empirically prove that our method is promising and data efficient for attentive reasoning and could successfully provide quality explanations and diverse rules for a variety e-commerce business tasks. 
\end{itemize}

%% file: preliminary.tex
\section{Preliminary}

\subsection{Knowledge Graph Embedding}
\begin{defn}
(Knowledge Graphs) A knowledge graph $\mathcal{G} = \{\mathcal{E}, \mathcal{R}, \mathcal{T}\}$, where $\mathcal{E}$, $\mathcal{R}$ and $\mathcal{T}$ are the set of entities, relations and triples. 
A triple $(h,r,t) \in \mathcal{T}$ includes a head entity $h \in \mathcal{E}$, a relation $r \in \mathcal{R}$ and a tail entity $t \in \mathcal{E}$.
\end{defn}
% knowledge graph embedding
KGE methods represent entities and relations in continuous vector spaces as vectors or matrices, called \emph{embeddings}, which are typically denoted by bold character $\mathbf{h}, \mathbf{r}$ and $\mathbf{t}$.
TransE~\cite{TransE:conf/nips/BordesUGWY13} is a typical KGE method which regards relation embedding $\mathbf{r}$ as a translation from the head entity embedding $\mathbf{h}$ to the tail entity embedding $\mathbf{t}$ in vector space and assumes that $\mathbf{h} + \mathbf{r} = \mathbf{t}$ for true triples. The score function in TransE is $f(h,r,t) = \|\mathbf{h} + \mathbf{t} - \mathbf{t}\|$ and assumes true triples get smaller scores and false triples get larger scores.

\subsection{Link Prediction and Attentive Reasoning}
\label{preliminary:LP}
Link prediction is an important knowledge graph reasoning task  defined as follows:
\begin{defn}
(Link Predictions in Knowledge Graphs) Given a knowledge graph $\mathcal{G} = \{ \mathcal{E}, \mathcal{R}, \mathcal{T}\}$, link prediction is to predict missing  triples given two elements of them, 
including tail entity prediction $(h, r,?)$, relation prediction $(h, ?, t)$, and head entity prediction $(?, r, t)$.
\end{defn}
General link prediction task does not distinguish the importance of relations and typically trains over all relations simultaneously. It is targeted to predict missing triples over all relations, while for attentive reasoning, good performance are concerned and expected mainly for target relations. Thus we propose partial link prediction as a task for attentive reasoning, defined as follows:
\begin{defn}
(Partial Link Prediction(PLP)) Given a KG $\mathcal{G} = \{ \mathcal{E}, \mathcal{R}, \mathcal{T}\}$ and a set of target relations $\mathcal{R}_{target} \in \mathcal{R}$, partial link prediction is to infer new triples $(h^\prime, r^\prime, t^\prime) \notin \mathcal{G}$ that $r^\prime \in \mathcal{R}_{target}$. 
\end{defn}

\subsection{Transferable rules}
\label{sec:Transferable rules}
Transferable rules play pivotal role in collecting high level business knowledge for reasoning task. In our system, over 3+ million rules have been collected to support daily business operation. An example rule is illustrated as below, 
\begin{equation}
(X, livesIn, Japan) \gets (X, hasWife, Z) \land (Z, livesIn, Japan)
\label{path-rule}
\end{equation}
indicating that couples usually lives in the same place. As shown in the example, rules are in the form of
	$head \gets body$, 
indicating that $head$ can be inferred from $body$. 
Both $head$ contains one atom and $body$ is composed with one or more atoms. An atom is a triple with one or two variables, denoted as uppercase characters (e.g. X, Y, Z).
Multiple atoms in rule body are organized with logical conjunctions. With all variables replaced in a rule, we called it a \textit{grounding} of this rule.  

If all variables in a rule appear twice in different atoms, it is a path rule, like (\ref{path-rule}).
Path rule reveals the relations between relations in KG. 
Association rules with specific relation and entity pairs are also common and important. For example,$(X, suitableFor, \text{Yoga}) \gets (X, isA, \text{Yoga Mat})$ which 
 reveals that if an item is a Yaga mat then it is suitable for Yoga. Association rule is especially common in e-commerce scenario but few methods are proposed for it in KG.

%% file: method.tex
\section{Method}
\label{sec:method}

\subsection{Attentive Reasoning Framework}
In order to make the method meet three desiderata in e-commerce application, the key point is to build correlation between triples. Thus in our method, \emph{the score function for $(h,r,t)$ is defined based on not only $\mathbf{h}, \mathbf{r}$ and $\mathbf{t}$, but also its weighted neighbor triples}. 

\begin{figure}[!htbp] 
\centering 
\vspace{-3mm}
\subfigure[1 layer]{
	\label{1-layer}
	\includegraphics[scale=0.16]{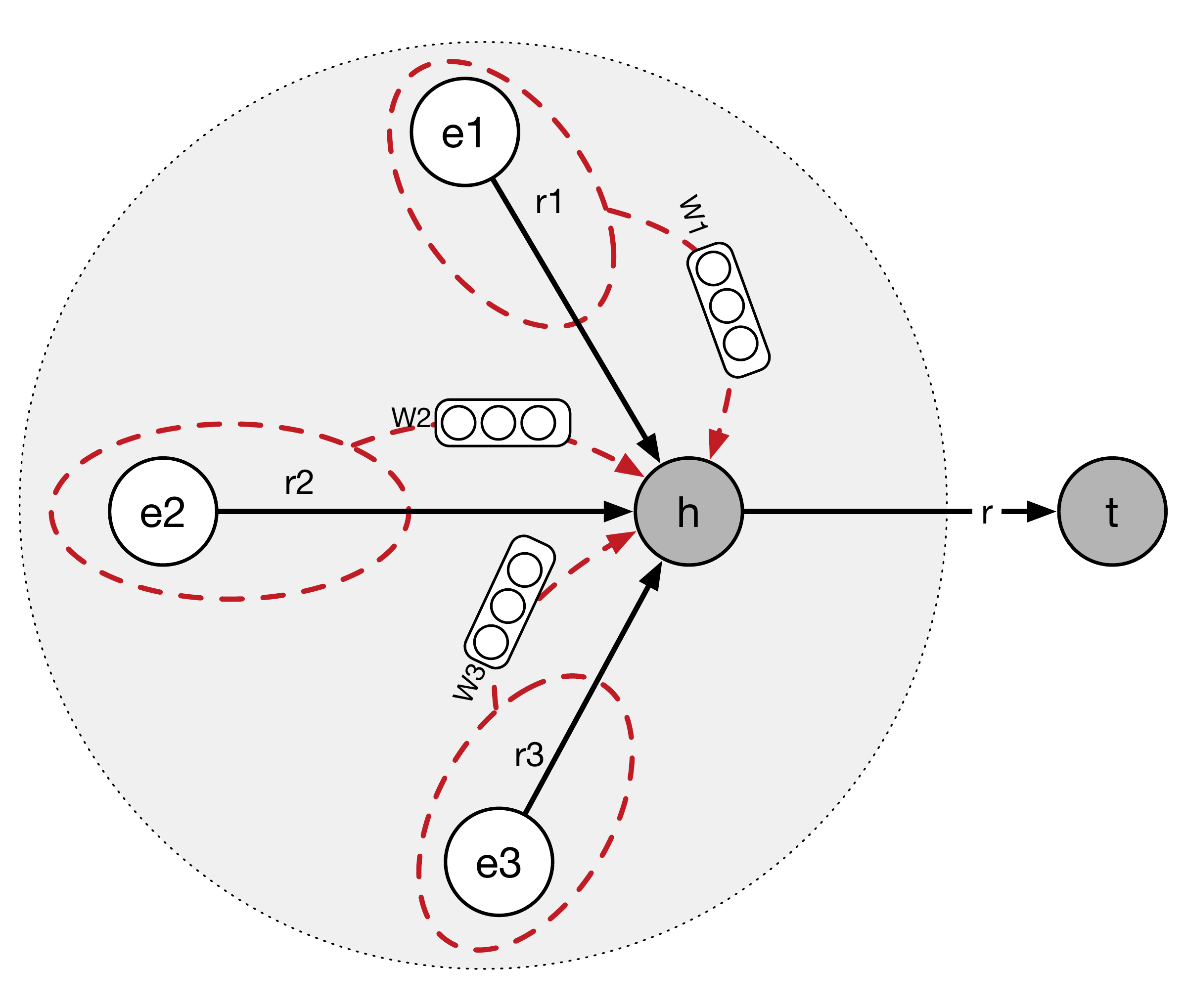}
	}
\subfigure[2 layers]{
	\label{2-layers}
	\includegraphics[scale=0.16]{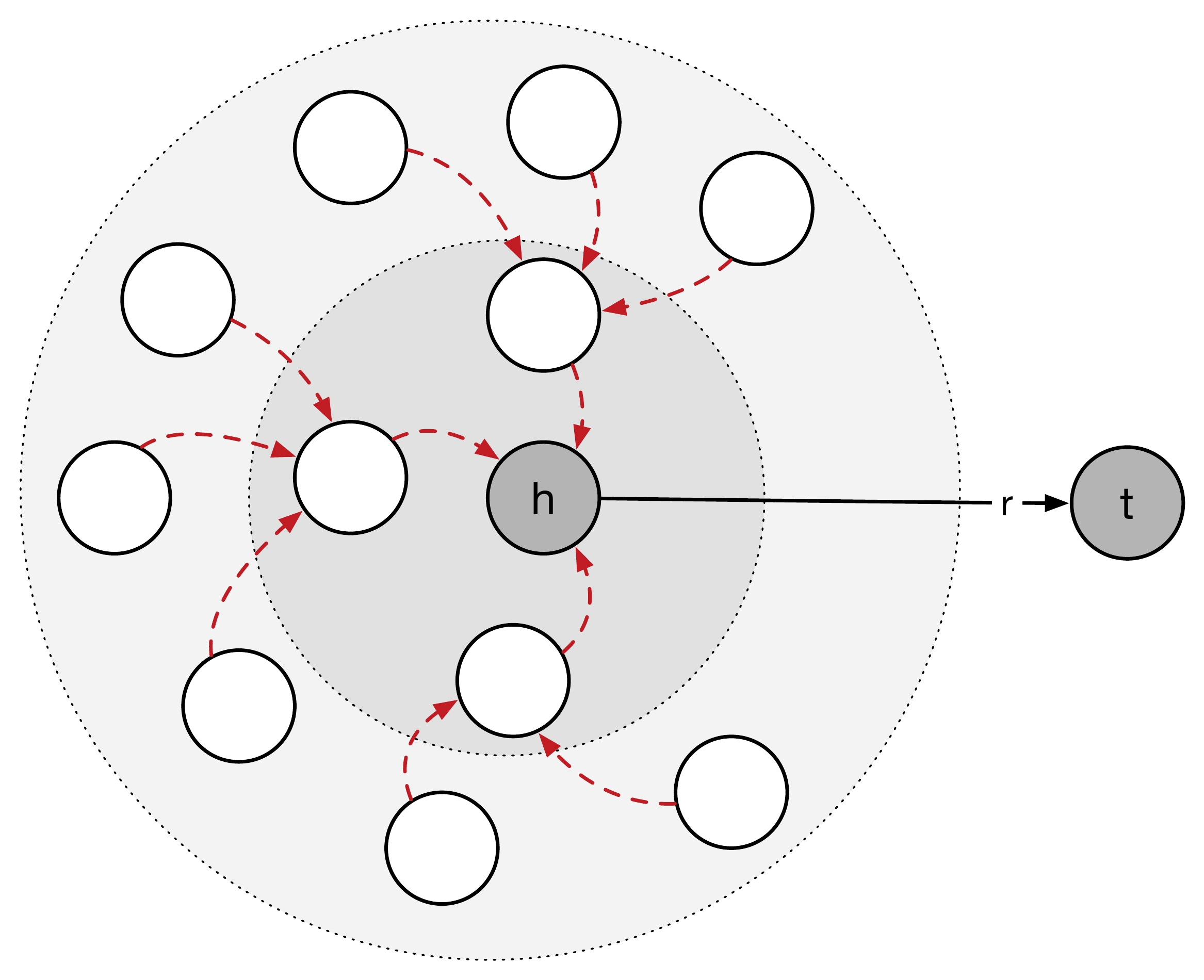}
}	
\vspace{-3mm}
\caption{General ideal of the explainable knowledge graph attention network.
}
\label{general-idea}
\vspace{-3mm}
\end{figure}

In this work, we define one-degree neighbor triple of $(e_1, r, e_2)$ as $(e^\prime, r^\prime, e_1)$ whose tail entity is $e_1$, and the common entity $e_1$  between these two triples is called \emph{shared entity}. 
Triples $(e^{\prime \prime}, r^{\prime \prime}, e^\prime)$ with $e^\prime$ as shared entity are one-degree neighbor triple of $(e^\prime, r^\prime, e_1)$ and  two-degree triples of $(e_1, r, e_2)$. Higher degree neighbor triples are defined in the same way. 
We devise a basic layer to aggregate triple's one-degree neighbor triple information before calculating their scores. The general idea of a basic layer is shown in Figure~\ref{general-idea}(a). For a target triple $(h,r,t)$ in one basic layer, we first calculate the shared entity representation in its one-degree neighbor triples based on entity and relation embeddings, and then conduct a weighted aggregation to update the representation for shared entity $h$. This makes head entity representation depends on one-neighbor triples. And we assume weights for different neighbor triples indicate their importance for current prediction. To enable the method sensible of higher degree neighbors, we stack multiple basic layers. Figure~\ref{general-idea}(b) shows how two layers works, we first aggregate shared entity representations from two-degree neighbors to one-degree neighbors and then aggregate shared entity representation from one-degree neighbors to $h$. In this way, the representation for $h$ will depend on all one-degree and two-degree neighbor triples.

\begin{figure*}[!htbp] 
\centering 
	\includegraphics[scale=0.43]{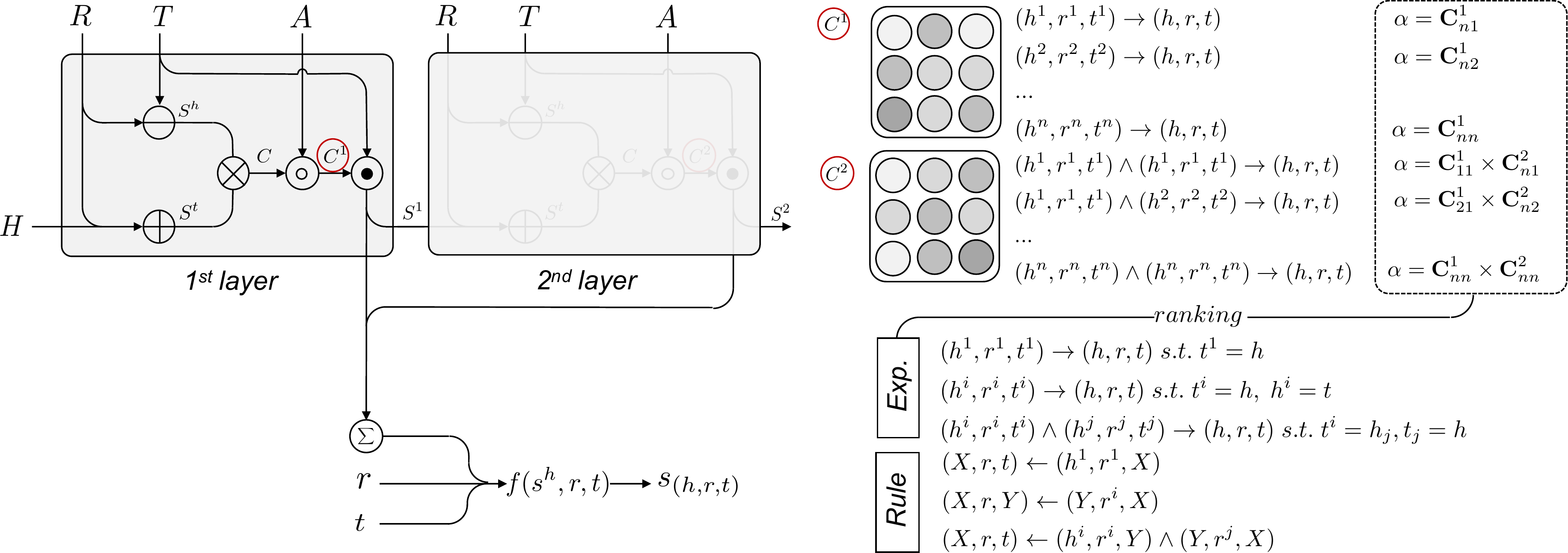}
	\vspace{-3mm}
	\caption{Generating Explanations and Transferable Rules.}
	\label{fig.method}
	\vspace{-5mm}
\end{figure*}

Like most KGE methods, we represent entities and relations as vectors and define a score function to model the truth value of triples based on translation assumption that $\mathbf{h} + \mathbf{r} = \mathbf{t}$ for a true triple $(h,r,t)$, 
where $\mathbf{h} \in \mathbb{R} ^d, \mathbf{r}\in \mathbb{R} ^d$ and $\mathbf{t}\in \mathbb{R} ^d$ and 
$d$ is embedding dimension.
For a training triple $(h,r,t)$, we input a neighbor subgraph containing its neighbor triples with in $k$ degree and itself. 
We use $N$ to denote the ordered set of triples in subgraph, where $N_i = (h_i, r_i, t_i)$ is the $i$th triple. We put the target triple at the end of $N$, thus $N_{n} = (h_{n}, r_{n}, t_{n}) = (h, r, t)$ where $n = |N|$.

\subsubsection{Basic layer.} 
With target triple $(h,r,t)$ and its neighbor triple set $N$. We first generate four matrices $\mathbf{H}, \mathbf{R}, \mathbf{T}$ and $\mathbf{A}$. 
 $\mathbf{H}\in \mathbb{R}^{n \times d}, \mathbf{R}\in \mathbb{R}^{n \times d}$ and $ \mathbf{T}\in \mathbb{R}^{n \times d}$ are the head entity, relation and tail entity matrix of $N$ with $\mathbf{H}_i = \mathbf{h}_i, \mathbf{R}_i = \mathbf{r}_i$ and $\mathbf{T}_i = \mathbf{t}_i$. 
 $\mathbf{A} \in \mathbb{R}^{n \times n}$ is the adjacent matrix for neighbor triple subgraph, where $\mathbf{A}_{ij} = 1$ if triple $N_j$ is one-degree neighbor of triple $N_i$, otherwise $\mathbf{A}_{ij} = 0$.

As each triple $N_i \in N$ could be one-degree neighbor triple of another one in $N$ with $t_i$ as shared entity, thus we first calculate shared tail entity representation for each triple $N_i \in N$
\begin{equation}
\mathbf{S}^t = \mathbf{H} + \mathbf{R}
\end{equation}
where $\mathbf{S}^t \in \mathbb{R}^{n \times d}$. $\mathbf{S}^t _i$ is shared tail entity representation for $N_i$ and we interpret it as what information could be transferred to $t_i$ if $N_i$ is one-degree neighbor triple of other triple $N_j$.

Similarly, each triple $N_i \in N$ might has one-degree neighbor triples in $N$ with $h_i$ as shared entity, thus we calculate shared head entity representation 
\begin{equation}
	\mathbf{S}^h = \mathbf{T} - \mathbf{R}
\end{equation}
where $\mathbf{S}^h \in \mathbb{R}^{n \times d}$.  $\mathbf{S}^h _i$ is shared head entity representation for $N_i$ and it could be interpreted as what information is expected to be transferred to $h_i$ from its one-degree neighbor triple.

With $\mathbf{S}^h$ and $\mathbf{S}^t$, we calculate similarity between each entity pair via dot product 
%Then we calculate similarity between $\mathbf{S}^h$ and $\mathbf{S}^t$ via 
\begin{equation} 
\mathbf{C} = \mathbf{S}^h (\mathbf{S}^t)^\top
\end{equation}
where $\mathbf{C} \in \mathbb{R}^{n \times n}$ is a similarity score matrix for different triple pair. $\mathbf{C}_{ij}$ is the similarity score between $N_i$ and $N_j$. The larger $\mathbf{C}_{ij}$ is, the more important $N_j$ is to $N_i$ as they get similar representation for their shared entity.  

$\mathbf{C}$ contains similarity score for each possible entity pair in the subgraph, while not all triples are neighbors. Thus we multiply $\mathbf{C}$ elementwisely with adjacent matrix $\mathbf{A} \in \mathbb{R} ^{n \times n}$ 
\begin{equation} 
\mathbf{C}^\prime = \mathbf{A} \circ \mathbf{C}
\end{equation} 
After multiplication, score $C_{ij}$ of triple pair $N_i$ and $N_j$ without shared entity will be $0$. 
Then we normalize each row of $\mathbf{C}^\prime$ via masked softmax function and get new normalized $\mathbf{C}^n$ with 
\begin{equation}
\label{equ:similarity matrix}
\mathbf{C}^n_{ij} = \frac{exp({\mathbf{C}^\prime _{ij}})}{\sum \limits_{\mathbf{C}^\prime _{ik} \ne 0} exp({\mathbf{C}^\prime _{ik}})}
\end{equation}
where $epx(x) = e^x$ and for all $i \in [1, n]$,  $\sum _{j=1} ^{n}\mathbf{C}^{n}_{ij} = 1$. $\mathbf{C}^{n}_{ij}$ could be interpreted as attentions for triple $N_j$ should $N_i$ pay for when aggregate shared entity information from its one-degree neighbors. 

With $\mathbf{C}^{n}$ and $\mathbf{S}^t$, we aggregated the shared entity representation of each $N_i \in N$ according to 
\begin{equation} 
\mathbf{S}^{+}  = \mathbf{C}^{n} \mathbf{S}^{t},\;\; \text{where}\;\; \mathbf{S}^+_{i} = \sum _{j = 1} ^n \mathbf{C}_{ij}^n \mathbf{S}_i ^t 
\end{equation}

Collecting all steps together, one basic layer function is
\begin{align}  
\mathbf{S}^{+} & = layer(\mathbf{A}, \mathbf{H}, \mathbf{R}, \mathbf{T}) \nonumber \\
& =  softmax(\mathbf{A} \circ (\mathbf{H} + \mathbf{R})(\mathbf{T} - \mathbf{R})^\top) \times (\mathbf{H} + \mathbf{R})
\end{align}
with one basic layer, we get the shared head entity representation for $(h,r,t)$ from its one-degree neighbor triples with $\mathbf{C}^n$ records the attention for each neighbor triple.

\subsubsection{Multiple Layers.} 
As introduced before, $\mathbf{S}^{+}$ from one layer is depended on triples' one-degree neighbors. In order to make the method sensible of higher degree neighbors, we make the basic stackable. 
To stack multiple layers, we input $\mathbf{S}^+$ as the head entity representation for $N$ instead of $\mathbf{H}$. 

Thus a $m$-layer model is
\begin{align}
	\mathbf{S}^1 &= layer(\mathbf{A}, \mathbf{H}, \mathbf{R}, \mathbf{T}, N) \\
	\mathbf{S}^2 &= layer(\mathbf{A}, \mathbf{S}^1, \mathbf{R}, \mathbf{T}, N) \\
	 & \qquad \qquad ... \\
	\mathbf{S}^m &= layer(\mathbf{A}, \mathbf{S}^{m-1}, \mathbf{R}, \mathbf{T}, N) 
\end{align}
where $\mathbf{S}^i \in \mathbb{R}^{n \times d}$ is the aggregated head entity representation $\mathbf{S}^+$ from $i$th layer. Within $\mathbf{S}^i$, the last row $\mathbf{S}_n^i$ is the representation for $h$ in target triple $(h,r,t)$.
With multiple $\mathbf{S}_n^i (i\in [1, m])$, we combine them together according to a weighted sum 
\begin{equation}
\label{omega}
 \mathbf{s}^h =   \sum _{k = 1} ^{m} \omega _k  \mathbf{S}^k_n, \; s.t.\; \sum_{k=1}^m \omega _k  = 1
 \end{equation}
where $\omega_k$ is a hyper-parameter denotes the weight for $k$th layer's representation during aggregation. $\mathbf{s}^h \in \mathbb{R}^{1 \times d}$ is the final representation for $h$, with which we calculate the  score of $(h,r,t)$ according to translation assumption 
\begin{equation}  s_{(h,r,t)} = \| \mathbf{s}^{h}  + \mathbf{r} - \mathbf{t} \|_{1/2} 
\label{score-function}
\end{equation}
where $\| \mathbf{x} \|_{1/2}$ is the L1 or L2 norm of vector $\mathbf{x}$. This equation evaluate how close the translated head entity representation is to tail entity embedding.
During training, we 
define a margin loss function to make the score of positive triple $s_{(h,r,t)}$ significantly smaller than its negative triple score  that 
\begin{equation}
L = \sum _{(h,r,t) \in \mathcal{T}} max(0, s_{(h,r,t)} + \gamma - s_{(h^\prime,r^\prime,t^\prime)})
\end{equation}
where $\mathcal{T}$ is the set of training triples. $s_{(h^\prime,r^\prime,t^\prime)} \notin \mathcal{T}$ is a negative triple for $(h,r,t)$ constructed by replacing $h$ or $t$ with another $e \in \mathcal{R}$.

\subsection{Generating Explanations and Rules}
\label{sec:explanation+rules}
In this section, we introduce how to generate explanations and rules. The overview of our method is shown in Figure~\ref{fig.method}.
\subsubsection{Generating explanations.}
In this work, we present explanations in the form similar to rules
$explanation \to prediction $, where
$prediction$ is the target triple and $explanation$ is composed by one or more triples. Thus an explanation can be expanded as 
\begin{align}
\label{explanation}
(h^1, r^1, t^1) &\land (h^2, r^2, t^2) \land ... \land (h^l, r^l, t^l) \to (h,r,t) \nonumber \\
 \quad \text{s.t.} \; &t^i = h^{i+1}, \; \forall i \in [1,l-1],\; t^l = h 
\vspace{-2mm}
\end{align} 
the expanded explanation means that $(h,r,t)$ is predicted because of triples $(h^1, r^1, t^1)$, $(h^2, r^2, t^2)$, ..., and $(h^l, r^l, t^l)$.  $l$ is the length of this explanation. We also add two constrains to make the order of triples unique and consider that a meaningful triple in $explanation$ should has shared entity with other triples. We constrain that $t^i = h^{i+1} (i \in [1,l-1])$ and $t^l = h$, thus they will form a path from $h^1$ to $t$. 
And if $h^1 = t$, they will form a closed circle. 

For each explanation, we assign a confidence score $\alpha \in [0, 1]$. And the higher the score is, the more confidence the explanation is. 
Score for explanation (\ref{explanation}) could be generated from our method with $l$ layers. 
The information of each triple $(h^i, r^i, t^i), i\in[1, l]$ is encoded by the $i$th layer in our method, and its correlation to the next triple is recorded in $\mathbf{C}^i$, the $i$th layer's $\mathbf{C}^n$. 
We calculate $\alpha$ for explanation (\ref{explanation}) via
\begin{equation}
\alpha =  \omega_l \times \mathbf{C}_{\hat{2}\hat{1}}^1 \times \mathbf{C}_{\hat{3}\hat{2}}^2 \times ... \times \mathbf{C}_{\hat{l}\widehat{l-1}}^{l-1} \times \mathbf{C}_{n \hat{l}}^l
\end{equation}
where $\hat{i}$ is the id of triple $N^i$ in explanation. 
$\mathbf{C}^k_{\hat{i}\hat{j}}$ is the value in $i$the row and $j$th column in $\mathbf{C}^k$
and it indicates how much triple $N^j$ contributes to triple $N^i$ in the $k$th layer.
$\omega _l$ is the weight for $l$ layers model, refer to Equation~(\ref{omega}).

For a $l$ layers model, there are $n + n^2 + ... + n^l$ possible explanations. During explanation generation, we calculate the confidence of each of them and output explanations with top $k$ scores. 

\subsubsection{Generating rules.}
We propose to generate rules from explanations. As we can see, the difference between an explanations and a rule is that explanation is composed by triples without variables while rule composed by atoms with variables.  Thus all explanations could be regarded as groundings of some rules.
As introduced in Section~\ref{sec:Transferable rules}, replacing variables in a rule with concrete entities will result in a grounding of it. Inversely, we could generate rules from explanations via replace concrete entities with variables. As a variable should be shared between atoms in a rule, thus replace shared entities in explanations with different variables. 
For explanation~(\ref{explanation}), one of the following rules could be generated
\begin{align}
	(V_1, r, t) \gets & ( V_{l-1} , r^l, V_l)\land (V_l, r^{l-1}, V_{l-1})\land... \nonumber \\
	&\land (V_1, r^2, V_2)\land  (h^1, r^1,V_1)\;\;  \text{if}\;  h \ne h^1\\
	(V_1, r, V_{l+1}) \gets & ( V_{l-1} , r^l, V_l)\land (V_l, r^{l-1}, V_{l-1})\land... \nonumber \\
	&\land (V_1, r^2, V_2)\land  (V_{l+1}, r^1,V_1)\;\;  \text{if}\;  h \ne h^1
\end{align}
where each $V_i$ is a variable. For one explanation, exact one rule will be generated. And different explanations might generate the same rule.
To control the quality of output rules, we select out rules  that are generated more than $\theta$ times from explanations, as the more frequently  a rule is generated, the more common the rule is.

%% file: experiment.tex
\section{Experiments and results}
\subsection{Dataset and Implementation Details}
\subsubsection{Dataset.} 

\input{tables/dataset.tex}
As shown in Figure~\ref{motivation}, the whole e-commerce KG is giant in volume containing over 70 billion triples and 3 million rules. In practice, we seldom use the whole KG for completion, and usually  select a subset of triples in a specific domain. Thus for demonstration purpose, experiments in this paper are also conducted on an e-commerce domain subset. This domain subset contains $789$ relations and  $72849$ entity in total, among which there are $11$ common and most important relations as target relations for attentive reasoning.

Among all triples, we randomly sample $20\%$ triples about the target relations into test set and $80\%$ for training. In test dataset, we only keep triples containing target relations.  
For partial link prediction task, there are two different dataset setting. 
The first one is PLP[all] in which all triples are included in training dataset, and the second one is PLP[part.] where only triples containing target triples are included in training dataset. We will train baselines on both these dataset and train our method on only PLP[part.].
For the case study of partial link prediction with only one target relation, the datasets are  PLP[life.] and PLP[cate.]. 
 
The detailed statistics of dataset for different experiment settings are shown in Table \ref{tab:dataets}. 

\input{tables/T-EPLP}

\input{./tables/T-prediction}

\subsubsection{Implementation details}
Before training, we introduce an inverse triple $(t, r^{-1}, h)$ for each $(h,r,t)$ in training dataset, where $r^{-1}$ is inverse relation of $r$. And according to translation assumption, we constrain $\mathbf{r}^{-1} = -\mathbf{r}$. 
Entity and relation embeddings are initialized from uniform distribution $U[-\frac{6}{\sqrt{d}}, \frac{6}{\sqrt{d}}]$  following~\cite{Init:journals/jmlr/GlorotB10}.

During training, we adopt mini-batch strategy with batch size set to $100$. 
For each batch, we sample triples from training dataset and construct a neighbor graph for them keeping neighbor triples within $2$ degrees. As the number of triples  increase dramatically when higher-degree neighbor triples are included and will bring more noise than valuable information, we filter $(h^\prime, r^\prime, t^\prime)$ if $h^\prime$ has more than $1000$ one-degree neighbors.
For all baselines and our proposal, embedding dimension $d=100$, margin $\lambda = 2$, and we generate one negative triple for each training triple via replacing entity with randomly sampled one from dataset. 
We optimize our method via Adaptive Moment Estimation(Adam) \cite{Adam:journals/corr/KingmaB14} with learning rate $0.0001$ and train it for max $5$ iterations with early stopping.

\subsection{Attentive Reasoning Evaluation}

\subsubsection{Evaluation metrics}
During test, for each test triple $(h,r,t)$, 
we generate multiple triple $(h,r,e)$ via replacing $t$ with each entity $e$ in the dataset, and calculate a score for them according to (\ref{score-function}). 
Then we rank these scores in ascend order and regard the rank of current test triple's score as its test rank. 
With a rank for each test triple, we calculate Mean Reciprocal Rank(\emph{MRR}) and the percentage of test triples that are ranked in top $k$(\emph{Hit@k(k={1,3,10}}) as evaluation metrics. 
Usually a better method should achieve  higher $MRR$ and $Hit@k$. 
Following previous KGE methods~\cite{TransH:conf/aaai/WangZFC14}, 
we also adopt textit{filter} settings during test during test that we filter replaced triples  existing in datasets before ranking. 

\subsubsection{Result analysis}
Partial link prediction results are shown in Table~\ref{tab:PLP}. 
Among all results presented, our method with both $1$ and $2$ layers with and without TransE initialization significantly outperform baselines. 
$1$ layer of our method initialized with TransE embedding achieves the best results on all metrics, outperforming  best results from baselines with $36.8\%, 16.7\%, 31.4\%$ and $54.5\%$ for MRR, Hit@10, Hit@3 and Hit@1 under \textit{filter} setting. 

In Table~\ref{tab:PLP}, all baselines trained with PLP[all] performs better than trained with PLP[part.], indicating that other triples do not contain target relations contribute to attentive reasoning. With all triples accessible during train, baselines could make prediction via global information encoded in entity and relation embeddings. 
Our method encoding local information from neighbor triples outperforms baselines trained with PLP[all], indicating that neighbor triples contain valuable local information for partial link prediction.  
Initializing entity and relation embeddings in our method with pre-trained TransE embeddings also help achieve performance gains, showing that combining local and global information helps in partial link prediction task. 

Thus we could conclude that generally our method is more capable of partial link prediction than baselines and successfully meet the first desiderata, attentive reasoning, in e-commerce application.

\subsubsection{Case study.}
We give two case study on task with only one target relation, because in e-commerce application, there are a lot of independent tasks concerning only one relation in KG. The two cases are  1) \emph{lifestyle Recommendation} to predict the lifestyle given an item and relation $suitableFor$ and 2) \emph{Automatic Categorization} is to predict the category given an item and relation $isA$.

During experiment, we train our method with $2$ layers on dataset PLP[life.] and PLP[cate.] in Table~\ref{tab:dataets} for lifestyle recommendation and automatically categorization respectively. 
For baselines, we train them with PLP[all.] making all information available to achieve better performance.     
In Table \ref{tab:res-EPLP}, we present the prediction results and the percentage of triple used for each task. Prediction results show that our method achieves slightly better results than baselines for lifestyle recommendation and comparable results for automatic categorization. 
With comparable performance, our method is super data efficient, it is trained with $2.23\%$ and $0.25\%$ triples of the whole dataset for two tasks respectively, while baseline method are train with $100\%$. 
This proves the success of our method to reasoning based on local information from neighbor triples. 

\input{tables/T-case-study}
\subsection{Explanation Evaluation}
\label{explanation evaluation}
\subsubsection{Evaluation metrics.}
We evaluate the quality of explanations via  $Recall$ and $AvgSupport$ following \cite{CrossE:conf/wsdm/ZhangPZBC19},
% from statistic perspective. 
and how much explanation really help experts check triples quicker.  
An explanation for a test triple is valid if and only if there exist similar structure as support in KG. 
For example, if there are triples \textit{(Item2, suitableFor, Middle Age)} and \textit{(Item2, collorStyle, Crew Neck)} then explanation 
\begin{equation}
\nonumber
	(Item1, colloarStyle, Crew\; Neck) \to (Item1, suitableFor, Middle Age) 
	\label{exp.explanation} 
\end{equation}
is a valid one. And we could frame these two supporting triples in the same form as explanation 
and it is regarded as a \textit{support} for above explanation. 
$Recall$ is the  percentage of test triples that the method can provide valid explanations for.
$AvgSupport$ is the average number of support can be found in training data for test triples with valid explanations. 

For our method, we generate explanations for each triple following section \ref{sec:explanation+rules} and output explanations within top $3$ from $1$ layer model with initialized with TransE which achieve the best results in Table~\ref{tab:PLP}.
For comparison, there is no existing explainable KGE method available. Thus  
we calculate similarity matrix $\mathbf{C}$ for TransE with its trained embeddings according to equation~(\ref{equ:similarity matrix}) and then generate explanations in the same way as our method. For justice,  we generate explanation for TransE with one-degree neighbor triples. 

\subsubsection{Results analysis.}
\begin{figure}[htbp] 
\centering 
\vspace{-5mm}
\subfigure[]{
	\label{relation-variation}
	\includegraphics[width=40mm,height=24mm]{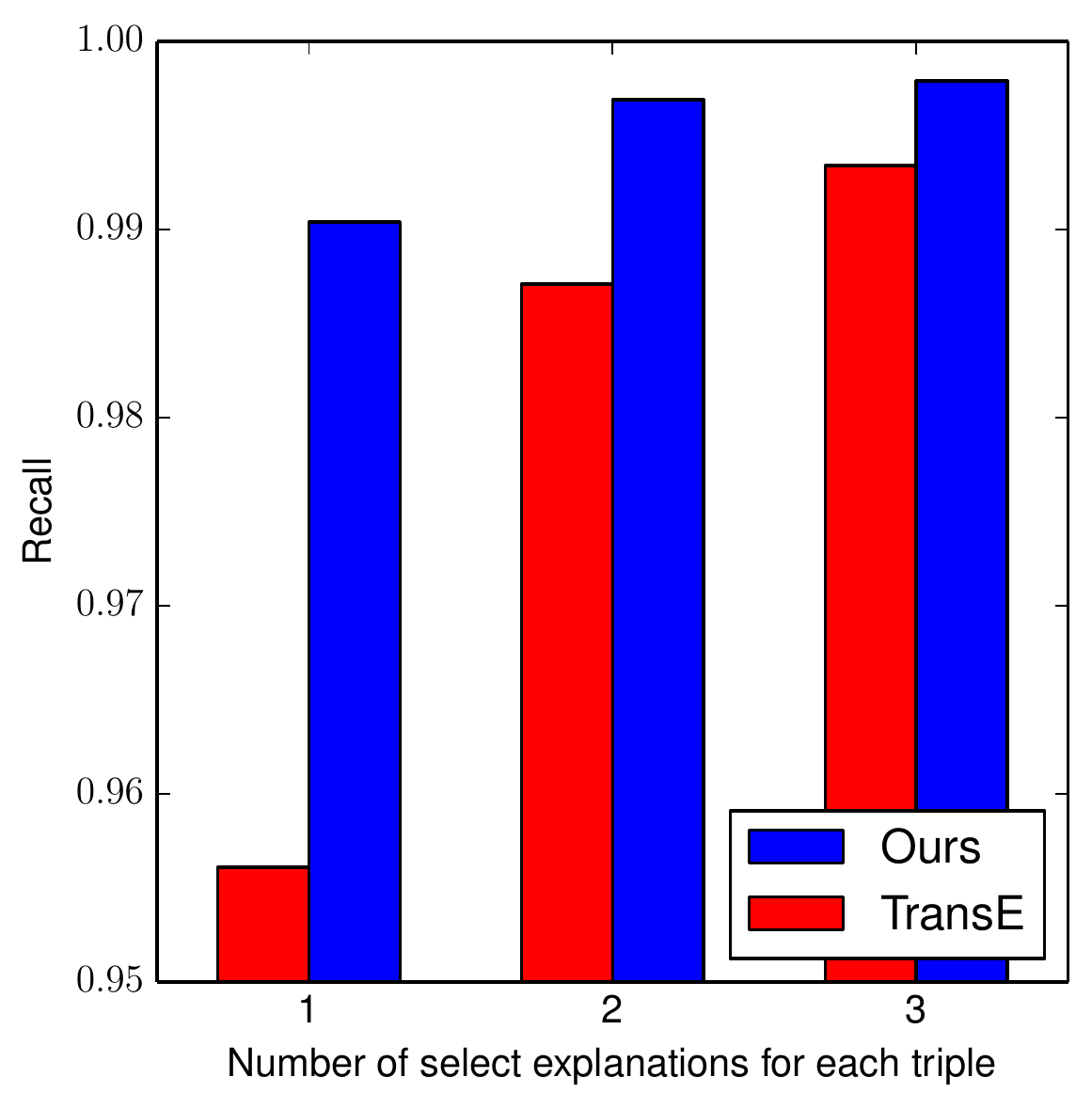}
	}
\subfigure[]{
	\label{HC}
	\includegraphics[width=40mm,height=24mm]{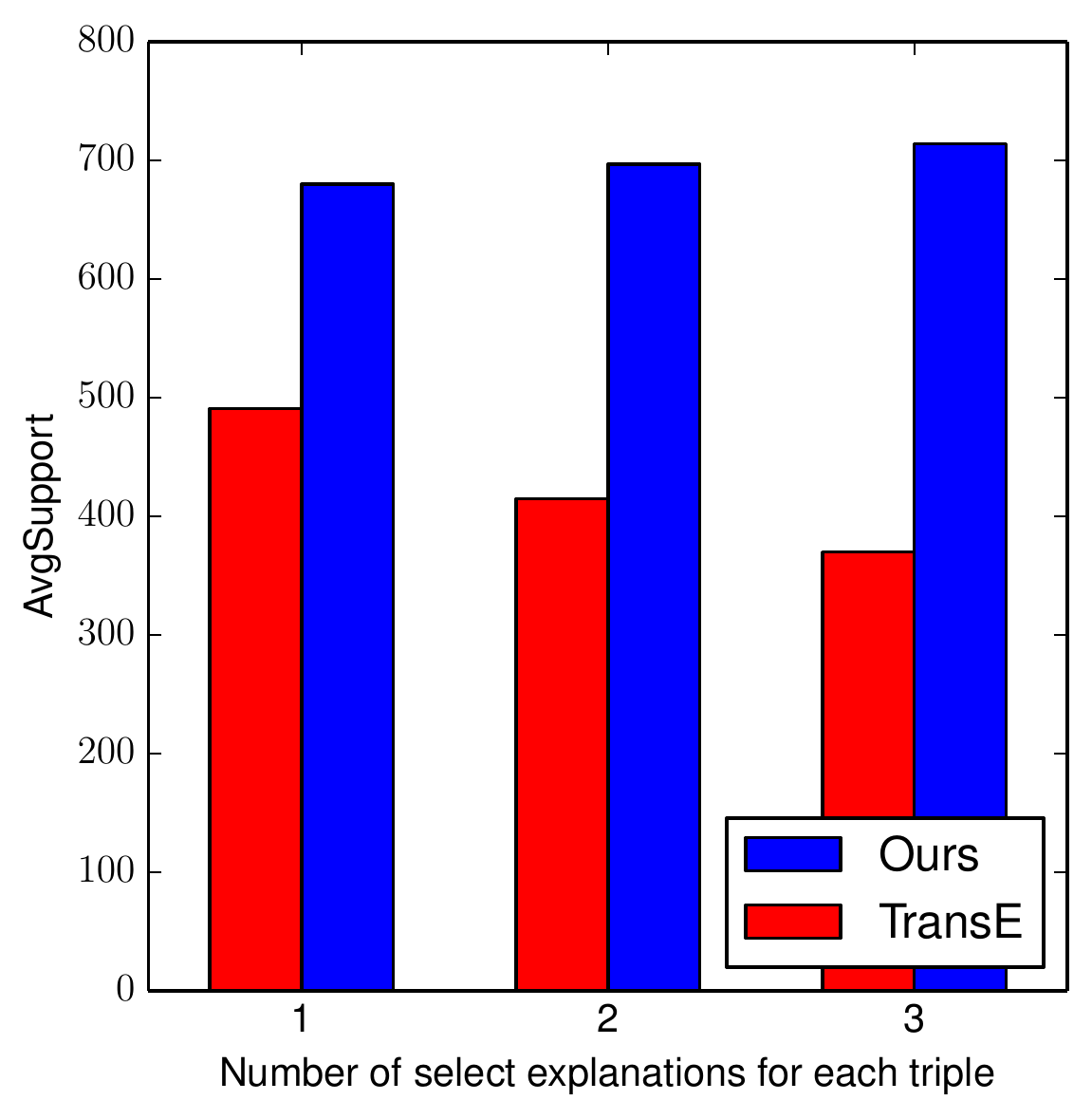}
}
\vspace{-3mm}
\caption{Results of explanation evaluation.}
\label{res-explanation}
\vspace{-5mm}
\end{figure}

In Figure~\ref{res-explanation},
From \textit{Recall}, we could see that  valid explanation could be provide for over $95\%$ test triples from both TransE and our method, and our achieve slightly higher \textit{Recall}. With $k$ increasing, Recall of both methods keep increasing. 
For AvgSupport, our method achieves significantly higher results than TransE, with average over $600$ supports for valid explanations. And with $k$ increasing from $1$ to $3$, AvgSupport from TransE keeps decreasing and ours keeps increasing, this indicates that top $3$ explanations from our method are with the same quality while in TransE, top $1$ explanation has higher quality than the second and third one. And generally, the quality of explanation from our method is better than that from TransE.

As the motivation of providing explanations from algorithms in our e-commerce application is to help experts check the correctness of predicted triples before added into KG, we preform a manual experiment to evaluate whether or not our method could improve the efficiency of human-machine collaboration cycle. In this experiment, we present a certain number of predicted triples to a group of e-Commerce experts to check their correctness, under settings with or without explanations. We then calculate and record the average precision and the average time (in seconds) they use to finish their jobs. The results are shown in Table~\ref{tab:explanation-manual}. 

\begin{table}[!hbpt]
	\centering
	\small
	\caption{Manually explanation evaluation.}
\vspace{-3mm}
	\begin{tabular}{l |c c }
	\toprule
	 & \textbf{Precision}  & \textbf{Time} \\
	 \hline
	 without explanation  & 69.33 &862.5(100\%) \\
	 explanation from TransE  & 64.33 &787.3(91.28\%)\\
	 explanation from ours  &69.67 &491.0(56.9\%)\\
	\toprule
	\end{tabular}
	\vspace{-5mm}
	\label{tab:explanation-manual}
\end{table}
Table~\ref{tab:explanation-manual} shows that provided with explanation from TransE, precision decreases a bit while with explanation from our method, the precision keeps comparable. This reveals that quality explanation is important. For the time cost for justification, the result reveals that explanation helps reduce the time cost, especially with our method that experts only need $56.9\%$ time for the same checking task compared with no explanation. Thus we could conclude that explanations from our method significantly improve the  efficiency of manually checking which is highly valuable in real life business.

\subsubsection{Case study.} 
Table~\ref{tab:explanation-example} shows three case studies of a test triple with top $3$ explanations from TransE and our method. 
In the first case, TransE focuses on key words from \texttt{Item1}s' title including \texttt{GrandFather}, \texttt{Middle-aged} and \texttt{Father} and our method make the prediction because normal clothes with crew neck and long sleeve are suitable for middle age. Both of them make sense and are easy to understand. 
In the second case, TransE still focuses on key words from title, while our method could pay attention to related properties. In the third case, the only convincing explanation is from our method that it predicts \texttt{Item3}'s brand is \texttt{Tianzi}, because its title include \texttt{Tianzi}.

\subsection{Rule Evaluation}
\subsubsection{Evaluation metrics.}
We generate rules from explanations for triples in both training and test dataset and calculate the number of rules, the number of high quality rules and the percentage of high quality rules in results, to evaluate the capability of generating rules. Usually, a better method should give higher number for all three metrics. Quality rules are regarded as rules generated at least  $5$ times among all triples, because different explanations might generate the same rule. 
%During rule generation, different explanations might generate the same rule. 
For example,  $exp_1$ that `(Item1, $suitableFor$, Middle Age)$\gets$ (Item1, $CollarStyle$, Crew Neck)' and $exp_2$ that `(Item2, $suitableFor$, Middle Age) $\gets$ (Item2, $CollarStyle$, Crew Neck)' will generate the same rule `(X,  $suitableFor$, Middle Age) $\gets$ (X, $CollarStyle$, Crew Neck)'. 
Rules generated with multiple times indicating they are common in the knowledge graph and are quality ones. 
Hight quality rules in this work are defined as rules with head coverage $HC>0.7$ following \cite{IterE:conf/www/ZhangPWCZZBC19} with an addition condition that it should has at least $20$ supports. Head coverage of a rule $head \gets body$ is defined as 
$ HC = \frac{\#(head \gets body)}{\# (head)}$.
$\# (head)$ is the number entities combination that satisfy the atoms in head atom and $\#(head \gets body)$ is the number of entities combination that satisfy the whole rule. And satisfying means the triples of replacing variables with entities all exist in knowledge graph. 

The most widely known rule mining method for knowledge graph is AMIE\footnote{Code from https://www.mpi-inf.mpg.de/departments/databases-and-information-systems/research/yago-naga/amie/}, thus we apply it as a baseline. During implementation, we constrain the length of rule body at most to $2$. We also generate rules from TransE's top $3$ explanations as well as ours.

\subsubsection{Results analysis.}
Table~\ref{tab:rule evaluation} shows results of rule evaluation. 
With AMIE, no quality rule is learned, because target path rules of AMIE is not quite common in our e-commerce knowledge graph. 
Generating rules from explanations shows promising results, with more than $5$ thousand quality rules and more than $1$ thousand high quality rules learnt. 
Among all evaluation metrics, our method outperforms significantly than TransE, with $589$ more quality rules, $593$ more high quality rules, and  $12.09\%$ more percent of high quality rules. 
To justify the usage of rules for knowledge graph completion,  
we also present how many new triples could be inferred from the hight quality rules. As shown in Table~\ref{tab:rule evaluation}, hight quality rules from TransE infers $50948$ triples and ours infer $163452$. Both are more than triples in test set. From our method, the number of new triples are comparable with the number of triples related to target relations in training set, indicating that e-commerce KG is far away from incomplete and rules are obvious valuable to KG completion.  

\subsubsection{Case study.}
In Table~\ref{tab:explanation-example}, we show a case study of possible rule generated from each explanation. With our method, exact one rule could generated from an explanation. 
Table~\ref{tab:explanation-example} shows that most of the rules are association rules describing the correlation between different item properties, and a few of them are path rules, like (\texttt{X}, \textit{bransIs}, \texttt{Y}) $\gets$ (\texttt{X}, $titleInclude$, \texttt{Y}). This shows the necessity compound types of rules in our application. 
 
\begin{table}
\centering
\small
\caption{Rule evaluation results. `(\%)' denotes the percentage of high quality rules in results. }
\vspace{-3mm}
	\begin{tabular}{l | c |c | c | c}
	\toprule
	     &\textbf{\# rules} & \textbf{\# HQr} & \textbf{(\%)} & \textbf{\# inferred triples}  \\
	 \midrule
	  AMIE &0 &0 & 0 & 0 \\
	 TransE & 8971 & 1411 & 16.06\% & 50948 \\
   	 \hline
	  \textbf{X-KGAT} &\textbf{9560} & \textbf{2004} & \textbf{31.42\%} & \textbf{163452} \\ 
	  \bottomrule
	\end{tabular}
\vspace{-5mm}
\label{tab:rule evaluation}
\end{table}

%% file: tables/dataset.tex
%\begin{table}[!hbpt]
%	\centering
%	\small
%	\caption{Statistics of e-commerce KG datasets. \textbf{\#E},\textbf{\#R}, \textbf{\#R$_{tar.}$} and \textbf{\# Cand.} are the number of entities, relations, target relations and candidates.}
%	\begin{tabular}{l c c c c  c c}
%	\toprule
%	  & \textbf{\#E} & \textbf{\#R} & \textbf{\#R$_{tar.}$} &\textbf{\#Train} & \textbf{\#Test} & \textbf{\# Cand.}\\
%	\hline
%	\textbf{PLP[all]} &72849 & 789 & 11 &1789717 & 45520 & 72849\\
%	\textbf{PLP[part.]} & 72849 & 789 & 11 & 183914 & 45520 & 72849\\
%	\textbf{EPLP[life.]} & 72849 & 789 & 1& 40049 & 9951 & 10 \\
%	\textbf{EPLP[cate.]} & 72849 & 789 &1 & 4550 & 1163 & 86 \\
%	\bottomrule
%	\end{tabular}
%	\label{tab:dataets}
%\end{table}

%\begin{table}[!hbpt]
%	\centering
%	\small
%	\caption{Statistics of datasets for different experiment setting.  \textbf{\#R$_{tar.}$} and \textbf{\# Cand.} are the number of target relations and candidate entities in the task. \#Train and \#Test are the number fo triples in training and test dataset.}
%	\begin{tabular}{l c c c c }
%	\toprule
%	  &  \textbf{\#R$_{tar.}$} & \textbf{\# Cand.} &\textbf{\#Train} & \textbf{\#Test} \\
%	\hline
%	\textbf{PLP[all]} & 11 & 72849 &1789717 & 45520 \\
%	\textbf{PLP[part.]} & 11 & 72849 & 183914 & 45520 \\
%	\textbf{PLP[life.]} & 1 & 10 & 40049 & 9951 \\
%	\textbf{PLP[cate.]} &1 & 86 & 4550 & 1163  \\
%	\bottomrule
%	\end{tabular}
%	\label{tab:dataets}
%\end{table}

\begin{table}[!hbpt]
	\centering
	\small
	\vspace{-5mm}
	\caption{The subset of e-Commerce KG. \textbf{\#E},\textbf{\#R}, and \textbf{\#R$_{tar.}$} are the number of entities, relations, and target relations.}
	\vspace{-3mm}
	\begin{tabular}{l c c c  c c}
	\toprule
	  & \textbf{\#E} & \textbf{\#R} & \textbf{\#R$_{tar.}$} &\textbf{\#Train} & \textbf{\#Test} \\
	\hline
	\textbf{PLP[all]} &72849 & 789 & 11 &1789717 & 45520 \\
	\textbf{PLP[part.]} & 72849 & 789 & 11 & 183914 & 45520 \\
	\textbf{PLP[life.]} & 72849 & 789 & 1& 40049 & 9951  \\
	\textbf{PLP[cate.]} & 72849 & 789 &1 & 4550 & 1163  \\
	\bottomrule
	\end{tabular}
	\label{tab:dataets}
	\vspace{-3mm}
\end{table}

%% file: tables/T-EPLP.tex
%\begin{table}
%	\centering
%	\small
%	\caption{Results of two case study of partial link prediction.}
%	% \vspace{-3mm}
%	\begin{tabular}{l| c | c| c| c| c| c}
%	\toprule 
%	 \multirow{2}{*}{\textbf{Method}}& \multicolumn{3}{c|}{\textbf{Lifestyle Recommendation}} & \multicolumn{3}{c}{\textbf{Automatic Categorization}} \\
%	 \cline{2-7}
%	 & \textbf{Hit@1} & \textbf{Hit@3} & \textbf{Hit@5} & \textbf{Hit@1} & \textbf{Hit@3} & \textbf{Hit@5} \\
%	\hline
%	TransE & 0.5404 & 0.7517 & 0.8433 & 0.2527 & 0.4462 & 0.5425 \\
%	TransH & 0.8689 & 0.9557 & 0.9830 & 0.8357 & 0.9587 & 0.9871 \\
%	TransR & 0.8418 & 0.9846 & 0.9948 & 0.8908  & 0.9750 & 0.9871 \\
%	DistMult&0.7043 & 0.9890 & 0.9972 & 0.8521 & 0.9664 & 0.9879\\
%	\hline
%	\textbf{Ours} & 0.8680 & 0.9900 &0.9980 & 0.8541 & 0.9440 & 0.9700 \\
%	\bottomrule
%	\end{tabular}
%	% \vspace{-5mm}
%	\label{tab:res-EPLP}
%\end{table}

\begin{table*}
	\centering
	\small
	\caption{Results of two case study of partial link prediction.}
	\vspace{-3mm}
	\begin{tabular}{l| c | c| c| c| c| c| c| c }
	\toprule 
	 \multirow{2}{*}{\textbf{Method}}& \multicolumn{4}{c|}{\textbf{Lifestyle Recommendation}} & \multicolumn{4}{c}{\textbf{Automatic Categorization}} \\
	 \cline{2-9}
	 & \textbf{Hit@1} & \textbf{Hit@3} & \textbf{Hit@5} & \textbf{\# train triples(\%)} &\textbf{Hit@1} & \textbf{Hit@3} & \textbf{Hit@5} &\textbf{\# train triples(\%)}\\
	\hline
	TransE & 0.5404 & 0.7517 & 0.8433  & 1789717 (100\%)
	& 0.2527 & 0.4462 & 0.5425  & 1789717(100\%) \\
	TransH & 0.8689 & 0.9557 & 0.9830  & 1789717 (100\%)
	& 0.8357 & 0.9587 & 0.9871 & 1789717(100\%)\\
	TransR & 0.8418 & 0.9846 & 0.9948 & 1789717 (100\%)
	& 0.8908  & 0.9750 & 0.9871 & 1789717(100\%)\\
	DistMult&0.7043 & 0.9890 & 0.9972 & 1789717 (100\%)
	& 0.8521 & 0.9664 & 0.9879 & 1789717(100\%) \\
	\hline
	\textbf{Ours} & 0.8680 & 0.9900 &0.9980 & 40049 (2.24\%)
	& 0.8541 & 0.9440 & 0.9700 & 1163 (0.25\%) \\
	\bottomrule
	\end{tabular}
% 	\vspace{-3mm}
	\label{tab:res-EPLP}
\end{table*}

%% file: tables/T-prediction.tex
\begin{table}
	\centering
	\small
	\caption{Partial link prediction results. 
% 	Result shown in on cell is in the form of \textit{raw/filter}.
	Baseline results with `[all]' and `[part.]' are trained on dataset PLP[all] and PLP[part.] respectively. \textbf{Ours}-1 and \textbf{Ours}-2 denotes our method with $1$ and $2$ layers respectively. `(init)' denotes our method is initialized with TransE's embeddings. Bold numbers are the best results and underlined are the second.}
	\vspace{-3mm}
	\begin{tabular}{r| c | c| c| c}
		\toprule
		  \multirow{1}{*}{\textbf{Method}} & 
		\multicolumn{1}{c|}{\textbf{MRR}} & 
%		\multicolumn{2}{c}{\textbf{MR}} & 
		\multicolumn{1}{c|}{\textbf{Hit@10}} & \multicolumn{1}{c|}{\textbf{Hit@3}} & \multicolumn{1}{c}{\textbf{Hit@1}} \\
% 		\cline{2-9}
% 		 \textit{filter} & \textit{filter} & \textit{filter}  & \textit{filter} \\
		 \midrule
		 TransE[part.]  
% 		 &0.2901 
		 & 0.2921 
%		 &588.90 & 588.86  
% 		 &0.7250 
		 & 0.7353 
% 		 &0.5212 
		 & 0.5252 
% 		 &0.0097 
		 & 0.0111  \\
		TransH[part.]  
% 		&0.4228 
		&0.4298 
%		&101.18 &101.14  
% 		&0.6721 
		& 0.6724 
% 		&0.4681 
		&0.4722 
% 		&0.3060 
		& 0.3159 \\ 
		TransR[part.]  
% 		&0.1931 
		& 0.1975 
%		&345.13 & 345.09  
% 		&0.4542 
		& 0.4549 
% 		&0.2677 
		& 0.2717 
% 		&0.0493 
		& 0.0551  \\
		DistMult[part.]  
% 		&0.3597 
		& 0.3641 
%		&314.67/314.64 
% 		 &0.7329 
		 & 0.7246 
% 		 &0.3998 
		 & 0.4034 
% 		 &0.2094 
		 & 0.2155  \\
		\midrule
		 TransE[all]  
% 		 & 0.3925 
		 & 0.3935
%		 &  570.95/570.92 
		 
% 		 &0.7801 
		 & 0.7804  
% 		 &0.5342 
		 & 0.5364 
% 		 & 0.1847 
		 & 0.1851 \\
		 TransH[all]  
% 		 & 0.5316 
		 & 0.5360
%		 & \textbf{46.32/46.29} 
		 
% 		 & 0.7782 
		 & 0.7789 
% 		 & \underline{0.6160} 
		 & \underline{0.6188} 
% 		 & 0.3951 
		 & 0.4017 \\	 
		 TransR[all]   
% 		 &\underline{0.5352} 
		 & \underline{0.5400} 
%		 &\underline{62.81/62.77} 
		 
% 		 &0.7825 
		 & 0.7831 
% 		 & 0.6085 
		 & 0.6118 
% 		 &\underline{0.4069} 
		 & \underline{0.4139}  \\
		 DistMult[all]  
% 		 &0.5217 
		 & 0.5247 
%		 &101.24/101.21 
		 
% 		 &\underline{0.7897} 
		 & \underline{0.7903} 
% 		 &0.5869 
		 & 0.5902 
% 		 &0.3899 
		 & 0.3934 \\
		 \midrule	 		
		 \textbf{Ours}-1   
% 		 &0.6816 
		 & 0.6837 
%		 &77.05/77.03 
		 
% 		 &0.8868 
		 & 0.8870 
% 		 &0.7589 
		 & 0.7612 
% 		 &0.5710 
		 & 0.5737 \\
		 \textbf{Ours}-2 
% 		 &0.6727 
		 & 0.6748 
%		 &91.02/91.01 
		 
% 		 &0.9130 
		 & 0.9133 
% 		 &0.7692 
		 & 0.7712 
% 		 &0.5433 
		 & 0.5459 \\
		 \textbf{Ours}-1(init) 
% 		 &\textbf{0.7360} 
		 &\textbf{0.7388} 
%		 &148.12/148.10 
		 
% 		 &\textbf{0.9223} 
		 &\textbf{0.9223} 
% 		 &\textbf{0.8117} 
		 & \textbf{0.8133}  
% 		 &\textbf{0.6357} 
		 & \textbf{0.6393}\\
		 \textbf{Ours}-2(init) 
% 		 & 0.7118 
		 &0.7142 
% 		 &0.8990 
		 &0.8998 
% 		 &0.7901 
		 &0.7916 
% 		 &0.6096 
		 &0.6132\\
		\bottomrule
	\end{tabular}
	\label{tab:PLP}
	\vspace{-5mm}
\end{table}

%% file: tables/T-case-study.tex
\begin{table*}[!htbp]
\centering
\small
\caption{Three case study of explanations. The cell with `Invalid' means that selected triple do not have support in training dataset and the explanation is not presented.}
\vspace{-3mm}
\label{tab:explanation-example}
	\begin{tabular}{ l |l |l}
	\toprule 
%	\textbf{Method} &
	 \textbf{TransE}-explanations 
%	& \textbf{DistMult} 
	& \textbf{Ours}-explanations 
	& \textbf{Ours}-rules from explanations\\
	\toprule 
	\multicolumn{3}{c}{(\textbf{\texttt{Item1}}, \textbf{\textit{suitableFor}}, \textbf{\texttt{Middle Age}})} \\
	\hline
%	\textbf{Explanation 1} &
	(\texttt{Item1}, $titleHas$, \texttt{GrandFather}) &
%	(\texttt{Item1}, $hasProperty$, \texttt{Collar Style}) &	 
	(\texttt{Item1}, $SleeveStyle$, \texttt{Normal}) & 
	(X, \textit{suitableFor}, \texttt{Middle Age}) $\gets$ (X, $SleeveStyle$, \texttt{Normal}) \\
	 	
%	\textbf{Explanation 2} &
		(\texttt{Item1}, $titleInclude$, \texttt{Middle-aged}) &
%		(\texttt{Item1}, $timeTo Market$, \texttt{2018}) & 
		(\texttt{Item1}, $CollarStyle$, \texttt{Crew Neck}) &
		(X, \textit{suitableFor}, \texttt{Middle Age}) $\gets$ (X, $CollarStyle$, \texttt{Crew Neck})
		\\
%	\textbf{Explanation 3} &
		(\texttt{Item1}, $titleInclude$,\texttt{Father}) &
%		(\texttt{Item1}, $hasProperty$, \texttt{Style}) &  
		(\texttt{Item1}, $lengthOfSleeve$, \texttt{Long}) &
		(X, \textit{suitableFor}, \texttt{Middle Age}) $\gets$ (\texttt{X}, $lengthOfSleeve$, \texttt{Long})\\
  	\hline
  	\multicolumn{3}{c}{(\textbf{\texttt{Item2}}, \textbf{\textit{category}}, \textbf{\texttt{Glass}})} \\
	\hline
%	\textbf{Explanation 1} &
		(\texttt{Item2}, $titleInclude$, \texttt{Lemon Water}) & 
%		(\texttt{Item2}, $volumn$, \texttt{100ML-200ML}) & 
		(\texttt{Item2}, $patternOfGlass$, \texttt{None}) & 
		(\texttt{X}, \textit{category}, \texttt{Glass}) $\gets$  (\texttt{X}, $patternOfGlass$, \texttt{None})\\
%	\textbf{Explanation 2} &
		(\texttt{item2}, $titleInclude$, \texttt{Fruit Juice}) &
%		(\texttt{item2}, $hasProperty$, \texttt{Number}) & 
		(\texttt{Item2}, $shape$, \texttt{Round}) &
		(\texttt{X}, \textit{category}, \texttt{Glass}) $\gets$  (\texttt{X}, $shape$, \texttt{Round})
		\\
%	\textbf{Explanation 3} & 
		(\texttt{Item2}, $titleInclude$, \texttt{Lemon Teacup}) &
%		(\texttt{item2}, $hasProperty$, \texttt{Brand}) &
		(\texttt{Item2}, $number$, \texttt{1}) &
		(\texttt{X}, \textit{category}, \texttt{Glass}) $\gets$  (\texttt{X}, $number$, \texttt{1})\\
	\hline
	\multicolumn{3}{c}{(\textbf{\texttt{Item3}}, \textbf{\textit{bransIs}}, \textbf{\texttt{Tianzi}})} \\
	\hline
%	\textbf{Explanation 1} &
		(\texttt{Item3}, $titleInclude$, \texttt{For Woman}) & 
%		None &
		(\texttt{Item3}, $titleInclude$, Tianzi) &
		(\texttt{X}, \textit{bransIs}, \texttt{Y}) $\gets$ (\texttt{X}, $titleInclude$, \texttt{Y})\\
%	\textbf{Explanation 2} &
		(\texttt{Item3}, $titleInclude$, \texttt{Brief}) & 
%		(\texttt{Item3}, $hasProperty$, \texttt{Color}) &
		(\texttt{Item3}, $padded$, \texttt{Not Padded})&
		(\texttt{X}, \textit{bransIs}, \texttt{Y}) $\gets$ (\texttt{X}, $padded$, \texttt{Not Padded}) \\
%	\textbf{Explanation 3} &
		Invalid &
%		(\texttt{Item3}, $hasProperty$, \texttt{Size}) &
		Invalid &
		Invalid \\
	\bottomrule
	\end{tabular}
\vspace{-2mm}
\end{table*}

%% file: related-work.tex
\section{Related works}
\textbf{Knowledge Graph Embedding.}
Following TransE, there are a lot of other knowledge graph embedding methods proposed following TransE to improve its reasoning capability including TransH\cite{TransH:conf/aaai/WangZFC14}, TransR\cite{TransR:conf/aaai/LinLSLZ15} and TransD\cite{TransD:conf/acl/JiHXL015}. 
The vector space assumption of knowledge graph embedding is not unique. For example, DistMult~\cite{DistMul:conf/iclr/2015} adopts linear map assumption which assumes that, for a true triple $(h,r,t)$, $\mathbf{h}\mathbf{M}_r = \mathbf{t}$, where $\mathbf{M}_r$ is the matrix representation for relation $r$. ANALOGY~\cite{ANALOGY:conf/icml/LiuWY17} and ComplEx~\cite{ComplEx:conf/icml/TrouillonWRGB16} also following linear map assumption.
RotatE~\cite{RotatE} models relation as a rotation in complex vector space. 
Not all KGE methods are proposed with a certain vector space assumption, there are also methods based on neural networks. For example, ConvE\cite{ConvE:conf/aaai/DettmersMS018} and ConvKB\cite{ConvKB:conf/naacl/NguyenNNP2018} adopts a convolutional neural networks  to model the embeddings of triples. 
R-GCN\cite{R-GCN:conf/eswc/SchlichtkrullKBTW2018} and KBAT\cite{KBAT:/conf/acl/NathaniCSK2019} use Graph Convolutional Networks\cite{GCN:conf/iclr/KipfW2017} and Graph Attention Networks\cite{GAT:conf/lcir/VelivckovicCCRLB2018} considering the influence of n-hop neighborhood on the representation of a triple.
Further more, some methods incorporate not only triples but also additional information. Such additional information can be entity types in \cite{enttype1:conf/acl/GuoWWWG15, enttype2:conf/ijcai/XieLS16}, paths in \cite{path1:conf/emnlp/LinLLSRL15, path2:conf/emnlp/GuML15, path3:conf/acl/ToutanovaLYPQ16} or logical rules in \cite{rule1:conf/ijcai/WangWG15, rule2:conf/emnlp/GuoWWWG16, rule3:conf/naacl/RocktaschelSR15}.
A comprehensive survey of knowledge graph embedding can be found in \cite{kgembedding}.

\textbf{Graph Neural Network}(GNN) is a neural network-based approach that operates on graphs in non-Euclidean space. It is firstly proposed in \cite{gnn} which intends to learn the representation of each node that contains the information of neighborhood by GNN's propagation. Besides original GNN, some methods propose modifications of propagation. 
Convolutional operation is used in \cite{GCN1:journals/corr/BrunaZSL13, GCN2:journals/corr/HenaffBL15, GCN3:journals/corr/abs-0912-3848, GCN:conf/iclr/KipfW2017, graphsage}. Graph Convolutional Networks(GCN)\cite{GCN:conf/iclr/KipfW2017} is a typical approach that applies spectral convolutional operation on graph domain. Further more, GraphSage\cite{graphsage} propagate information by sampling and aggregate features from neighborhood and propose three aggregators. In Graph Attention Networks(GAT)\cite{GAT:conf/lcir/VelivckovicCCRLB2018} which applies attention mechanism on the step of propagation of GNN, the attention weight of aggregation from neighborhood to each node is calculated by self-attention method\cite{Transformer} and it overcomes the shortcoming of GCN which treats all neighborhood equally. There are also other modification like using LSTM\cite{LSTM} or GRU\cite{GRU} in propagation step, including gated graph neural networks\cite{GGNN} and Tree-LSTM\cite{GNN-Tree-LSTM}. \cite{gnn-review} gives a comprehensive review of graph neural networks. Our method incorporates the key ideal of GAT.

\textbf{Rule Learning.}
Many previous works have studied the mining method of different kinds of rules\cite{AMIE:conf/www/GalarragaTHS13, AMIE+, PRA:conf/emnlp/LaoMC11, SWARM, rulemine1:conf/semweb/HoSGKW18, RDF2Rules:journals/corr/WangL15i}. AIME\cite{AMIE:conf/www/GalarragaTHS13} and AMIE+\cite{AMIE+} mine Horn rules from knowledge graphs under the open-world assumption. Path Ranking Algorithm(PRA)\cite{PRA:conf/emnlp/LaoMC11} infers new facts for knowledge graphs by mining closed path rule. SWARM\cite{SWARM} proposed a method to mine semantic association rules by knowledge from instance-level and schema-level. 
Besides the above methods which search in knowledge graphs, there are also other approaches learning rules. For examples, NeuralLP\cite{NeuralLP} and DRUM~\cite{DRUM} are differentiable methods for learning logical rules, and DeepPath\cite{DeepPath} and MINERVA\cite{MINERVA:conf/iclr/DasDZVDKSM18} apply reinforcement learning methods to find reasoning paths as well as rules in KGs.